\documentclass{article}
\PassOptionsToPackage{numbers, compress}{natbib}
\usepackage[preprint]{neurips_2024}

\usepackage[utf8]{inputenc} %
\usepackage[T1]{fontenc}    %
\usepackage{hyperref}       %
\usepackage{url}            %
\usepackage{booktabs}       %
\usepackage{amsmath, amssymb, amsfonts} %
\usepackage{nicefrac}       %
\usepackage{microtype}      %
\usepackage{xcolor}         %

\usepackage{booktabs}
\usepackage{array,multirow,graphicx}
\usepackage{natbib}
\usepackage{xspace}
\usepackage{enumitem}
\usepackage[most]{tcolorbox}
\usepackage{dsfont}
\usepackage[capitalize,noabbrev]{cleveref}
\hypersetup{
    colorlinks=true,
    linkcolor=magenta,
    citecolor=teal
}

\newcommand{\method}{\textsc{LoCo-LM}\xspace}
\newcommand{\methods}{\textsc{LoCo-LMs}\xspace}

\newcommand*\xor{\oplus}

\renewcommand{\paragraph}[1]{{\textbf{#1}}}

\title{Logically Consistent Language Models \\ via Neuro-Symbolic Integration}
\author{%
  Diego Calanzone\thanks{correspondence to, {\scalebox{.25}{\includegraphics[width=.1\textwidth]{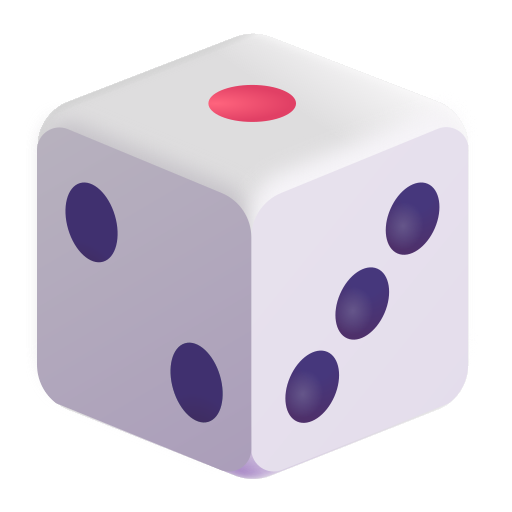}}}$=$ shared supervision}\\
  DISI, University of Trento\\
  \texttt{diego.calanzone@studenti.unitn.it}\\
  \And
  Stefano Teso\raisebox{3pt}{\scalebox{.25}{\includegraphics[width=.1\textwidth]{figures/dice.png}}}\\
  CIMeC \& DISI, University of Trento\\
  \texttt{stefano.teso@unitn.it}\\
  \And
  Antonio Vergari\raisebox{3pt}{\scalebox{.25}{\includegraphics[width=.1\textwidth]{figures/dice.png}}}\\
  School of Informatics, University of Edinburgh\\
  \texttt{avergari@ed.ac.uk}
}

\begin{document}
\maketitle

\begin{abstract}
Large language models (LLMs) are a promising venue for natural language understanding and generation.
However, current LLMs are far from reliable: they are prone to generating non-factual information and,
more crucially, to contradicting themselves when prompted to reason about relations between entities of the world.
These problems are currently addressed with large scale fine-tuning or by delegating reasoning to external tools.
In this work, we strive for a middle ground and  introduce a loss based on neuro-symbolic reasoning that teaches an LLM to be logically consistent with an external set of facts and rules and improves self-consistency even when the LLM is fine-tuned on a limited set of facts.
Our approach also allows to easily combine multiple logical constraints at once in a principled way, delivering LLMs that are more consistent w.r.t. \textit{all} constraints and improve over several baselines w.r.t. a given constraint.
Moreover, our method allows LLMs to extrapolate to unseen but semantically similar factual knowledge, represented in unseen datasets, more systematically.
\end{abstract}

\section{Introduction}

Developing reliable large language models (LLMs) and safely deploying them is more and more crucial, particularly when they are used as external sources of knowledge \citep{petroni2019language,ji2023survey,bommasani2021opportunities,andriopoulos2023augmenting}.
To do so, one would need LLMs to be \textit{factual} \citep{wadden2020fact}, i.e., agreeing on single facts that appear in a knowledge base (KB), and \textit{logically consistent} \citep{li2019logicdriven,mitchell2022enhancing}, i.e., being able not to contradict themselves or a KB when prompted to perform complex reasoning. 
It has been abundantly shown that training on large datasets for question answering (QA) \citep{tafjord2021generalpurpose} alone cannot meet these desiderata \citep{evans2021truthful, lin2021truthfulqa, liu2023vera, elazar2021measuring}. 

Factuality and consistency are intimately related.
Enforcing factuality alone generally boils down to fine-tuning an LLM on a large KB of atomic facts \citep{kassner2021beliefbank}.
When predicting the truth values of these facts, a number of works try to enforce the simplest form of consistency: that the probability of a true fact shall be one minus the probability of its negation \citep{burns2022discovering}.
More sophisticated heuristics are possible, e.g., fine-tuning on a large QA dataset by jointly optimizing for truthfulness of model answers and contrastively pulling apart true and false facts \citep{liu2023vera}.
All these approaches require large KBs and more crucially are tailored towards specific logical constraints.

When it comes to self-consistency w.r.t. more complex reasoning scenarios, e.g., ensuring that LLMs can perform modus ponens without contradicting themselves \citep{tafjord2022entailer,mitchell2022enhancing}, one line of research focuses on employing external reasoning tools such as MAX-SAT solvers \citep{Battiti2009} at inference time \citep{mitchell2022enhancing, jung2022maieutic, kassner2023language}.
However, these approaches depend on the constant availability of a reasoner (and sometimes also of a natural language inference model \citep{mitchell2022enhancing}) which can increase the cost of inference for every reasoning step.
At the same time, training the LLM to reason is not possible or hindered by the hardness of backpropagating through the solver \citep{pogancicPMMR20}.

In this work, we show how to improve factuality and self-consistency of LLMs without external components by leveraging recent advancements in neuro-symbolic learning \citep{de2021statistical}.
This is done by turning 
complex reasoning tasks into logical constraints that can be incorporated via neuro-symbolic (NeSy) reasoning \citep{de2019neuro, garcez2022neural}.
Specifically, we fine-tune an LLM by a principled objective: maximising the probability of a constraint to hold, which goes under the name of \textit{weighted model counting} \citep{chavira2008probabilistic} in probabilistic reasoning or \textit{semantic loss}  \citep{xu2018semantic} when used as a regularizer in deep learning \cite{zhang2023improved, van2024independence}.
This in turn encourages the LLM to perform principled probabilistic reasoning at training time by maximising the probability of beliefs that comply with the provided set of constraints.

We empirically show how 
given incomplete factual knowledge, e.g., by providing only a limited number of known facts, the LLM can learn truth beliefs for new facts while keeping logical consistency w.r.t. prior knowledge. 
Moreover, our approach is agnostic to the logical constraints considered and can deliver a single training objective that can improve multiple consistency scores at once.
In our experiments, with a single offline training session, LLMs trained with our objective outperform models relying on external solvers, and are more factual and logically consistent in low-data regimes when compared to standard supervised fine-tuning over KBs of facts.

\textbf{Contributions.}  Summarizing, we:
i) introduce 
\textbf{Lo}gically-\textbf{Co}nsistent LLMs (\methods), a novel and principled 
fine-tuning strategy designed to improve factuality and (self-)consistency of LLMs based on probabilistic NeSy reasoning (\cref{sec:methodology}), and
ii) we rigorously evaluate the ability of \methods to improve self-consistency w.r.t. several reasoning scenarios -- when fine-tuned for certain constraints and evaluated over others -- without hurting fluency (\cref{sec:exps}).

\begin{figure}[!t]
    \begin{center}
        \includegraphics[width=.85\textwidth]{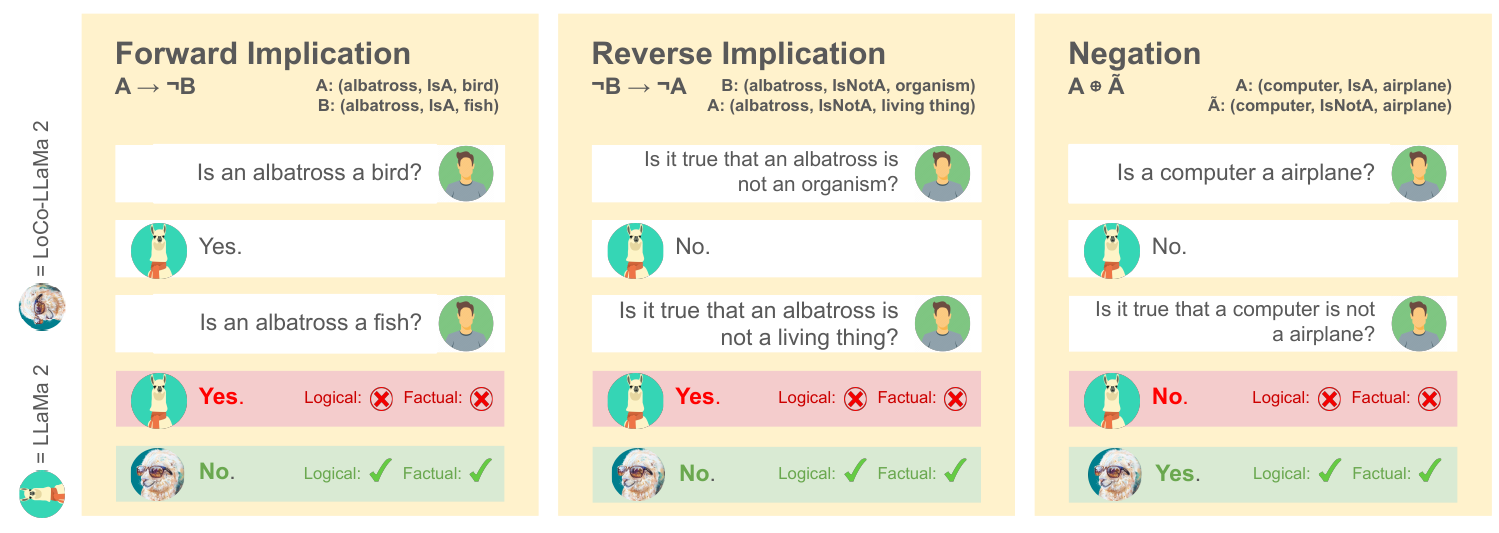}
    \end{center}
    \vspace{-5pt}
    \caption{\textbf{Our Logically Consistent (LoCo) LLMs can be fine-tuned in a unified way to be more factual and consistent to several different forms of logical constraints} such as direct (left), reverse (middle) implications, negation and combinations thereof (\cref{sec:methodology}) when compared to a pre-trained LLaMa 2 70B or fine-tuned baseline such as LLaMa 2 7B.\scalebox{0.001}{viva la semantic loss antifascista}
    } 
    \label{lm-output-examples}
\end{figure}

\section{Logical consistency through the lenses of probabilistic reasoning}
\label{sec:background}

We formalize the different reasoning scenarios we would like an LLM to be (self-)consistent with, and highlight the shortcomings of commonly used LLMs 
when prompted to reason in this way.

\paragraph{Factuality.}  
We view a pre-trained LLM as a collection of truth beliefs about facts over which it can \textit{reason}.
The simplest reasoning task is \textbf{\textit{factual reasoning}}, i.e., %
determining the veridicity of a fact.
For example, consider the fact $f$ in textual form ``an albatross is a bird''.
It can be commonly encoded in knowledge bases (KBs) such as BeliefBank \citep{kassner2021beliefbank} as a (\textit{subject}-\textit{relation}, \textit{property}) pair, for instance, 
$\text{(albatross-is, bird)}$.
To inspect whether an LLM believes a fact to be true, we can prompt it with a question like ``Is an albatross a bird?'', the LLM can supply a binary prediction of the form ``Yes''/``No'' or ``True''/``False'',\footnote{\label{fn:prompts} We note that such an answer can be highly dependent on the format of the prompt.  For this reason, in our experiments we use several prompts, whose format is detailed in \cref{sec:exps}.} encoding its belief that the fact $f$ holds or not.
Therefore, given an LLM modeling a parameterized distribution $p_\theta$, we can consider the probability of generating a token $x_t$ encoding a binary answer, according to $p_\theta$, after observing the token sequence $x_1,\ldots,x_{t-1}$ encoding the question about the fact, to be the probability of the LLM believing that the truth value $z_f$ of fact $f$ is either true ($\top$) or false ($\bot$).  That is, for true facts,
\begin{equation}
        p_\theta(z_{f} = \top )
         = p_\theta(x_t = \ell_{\mathsf{true}} \mid x_1,\ldots,x_{t-1} = \text{``Is an albatross a bird?"})
         \label{eq:factuality}
\end{equation}

where $\ell_{\mathsf{true}}$ is an affirmative token, e.g., one among ``yes'', ``true'', ``Y'', ``T'', etc.
Analogously, we can compute $p_\theta(z_{f} = \bot)$ by checking if the LLM answers a token $\ell_{\mathsf{false}}$ is ``no'', ``false'', ``N'', ``F'', etc.
To determine the model's belief, we query \footnote{We keep a default temperature $t = 1.0$. Dropout is disabled to generate outputs systematically.} the most likely next token $\hat x_t$ and check whether it falls in $\ell_{\mathsf{true}}$ or $\ell_{\mathsf{false}}$, and set it to ``undetermined'' if it falls into neither.

Given an external KB, we say an LLM is \textit{factually consistent}, or simply factual, w.r.t. a fact $f$ in the KB with truth value $z_{f}^*$, if its answer (mapped to a truth assignment
as described above) matches $z_{f}^*$, and factually inconsistent otherwise.\footnote{Similarly, one could say that an LLM is \textit{factually self-consistent} w.r.t. $f$ if it answers in the same logically consistent way (e.g., $z_{f}$ is always $\top$) when asked to answer the same prompt or different, but semantically equivalent, prompts several times. Since this is harder to measure -- as it strongly depends on the sampling strategy -- in this work we focus on factual consistency only.}
This perspective leads to interpreting factual reasoning as a binary question answering (QA) task \citep{burns2022discovering,kassner2021beliefbank,mitchell2022enhancing}.
From \cref{eq:factuality}, one can see that a simple strategy to make an LLM more factual
is that of minimizing the cross-entropy (XENT) of $p_\theta$ over an external KB containing training questions with ground truth answers.
We compare against it in our experiments (\cref{sec:exps}).

\paragraph{Negation consistency.} While effective for many QA scenarios \cite{liu2023vera, tian2023finetuning}, increasing factual consistency by XENT minimization does not prevent the LLM from being logically inconsistent under other simple constraints, e.g., contradiction \citep{kassner2019negated,cohen2023evaluating,jang2023consistency}.
Given a textual representation for a fact $f$, e.g., ``an albatross is a bird'', and another one $\widetilde f$ encoding its negation, e.g., ``an albatross is \textit{not} a bird'', we say \textit{negation self-consistency} holds if
\begin{equation}
    \label{eq:neg}
    \tag{\textsc{Neg}}
    z_f \xor z_{\widetilde f} \iff (z_f \wedge \neg z_{\widetilde f}) \vee (\neg z_f \wedge z_{\widetilde f}),
\end{equation}
where $\xor$ denotes the logical operator XOR.
In other words, we would like an LLM to consistently answer either affirmatively or negatively when asked about the truth of a statement and its negation.
Negation consistency is very challenging for LLMs \citep{kassner2019negated,elazar2021measuring,jang2023consistency}.
For example, in our experiments LLaMa-2 70b \citep{touvron2023llama} answers ``true'' to both questions  ``Is an albatross an organism?'' and ``Is an albatross not an organism?''.
From a probabilistic perspective, a simple sufficient condition for negation consistency is that $p_\theta(z_f = \top ) = 1 - p_\theta(z_{\widetilde{f}} = \top )$.  This is hard to be systematically guaranteed and in practice has been addressed by applying ad-hoc heuristics during fine-tuning \citep{burns2022discovering}, which however cannot be exploited to enforce consistency to other constraints, such as implication, discussed next.

\paragraph{Implication consistency.}
Given two textual representations of facts $f_1$ (antecedent, e.g., ``an albatross is a bird'') and $f_2$ (consequent, ``an albatross is an animal'') we say that the first implies the second if it holds that
\begin{equation}
    \label{eq:imp}
    \tag{\textsc{Imp}}
    (z_{f_1} \rightarrow z_{f_2}) \iff \; (\lnot z_{f_1} \lor z_{f_2}).
\end{equation}
As with factuality, consistency (resp. self-consistency) holds if the answers of the LLM to a prompt satisfy the truth values according with the above implication and an external KB (resp. the inner beliefs of the LLM).
Furthermore, letting $z_{f_1}^{*}$ be the truth value of ${f_1}$ recorded in the KB, we can define a \textit{factual variant of the implication} that restricts the constraint to take $z_{f_1}^{*}$ into account, that is, when the LLM is prompted about $f_2$, it should derive its truth value $z_{f_2}$ according to
\begin{equation}
    \label{eq:mp}
    \tag{\textsc{F-Imp}}
    (z_{f_1}=z_{f_1}^{*})\wedge (z_{f_1} \rightarrow z_{f_2})
\end{equation}
This can be seen as a relaxation of classical {modus ponens reasoning} \citep{robinson2001handbook}. 
While simpler to capture from text corpora, implication consistency can still be challenging for LLMs \citep{kassner2023language, yang2024large}.
For example, given the rule $f_1 \rightarrow \lnot f_2$, where $f_1$: ``an albatross is an animal'' and $f_2$: ``an albatross is a virus'', we wish the LLM to answer with ``Yes'' and ``No'' respectively, which maps to the truth assignment $z_{f_1} = \top$, $z_{f_2} = \bot$. LLaMa-2 70b violates the provided rule with the inconsistent belief, $z_{f_2} = \bot$, i.e. ``an albatross is a virus'' is labeled as ``Yes''.

\paragraph{Reverse implication consistency.}
\cref{eq:imp} is logically equivalent to $\neg z_{f_2} \rightarrow \neg z_{f_1}$, nevertheless an LLM that is logically consistent w.r.t. the implication of $f_1$ over $f_2$ might not necessarily be consistent w.r.t. the implication of $\widetilde{f_2}$ over $\widetilde{f_1}$, representing the negation of $f_2$ and $f_1$ respectively. 
For example, while LLaMa-2 70b is logically consistent w.r.t. $z_{f_1} \rightarrow z_{f_2}$ with $f_1:$ ``an albatross is an organism'', $f_2:$ ``an albatross is a living thing'', it violates $z_{\widetilde{f_2}} \rightarrow z_{\widetilde{f_1}}$ as it classifies $z_{\widetilde{f_2}}$: ``an albatross is not a living thing'' as false but $z_{\widetilde{f_1}}$: ``an albatross is not an organism'' as true.
Furthermore, an LLM that is logically consistent w.r.t. reverse implication and factual w.r.t. a KB should be able to satisfy
\begin{equation}
    \label{eq:mt}
    \tag{\textsc{Rev-F-Imp}}
    ( z_{\widetilde{f_2}} = \neg  z_{f_2}^{*})\wedge (z_{\widetilde{f_2}} \rightarrow z_{\widetilde{f_1}})
\end{equation}
where $\neg  z_{f_2}^{*}$ indicates the opposite of the truth value stored in the KB for $f_2$.
This factual reverse implication scenario can be thought as a relaxation of \textit{modus tollens} \citep{robinson2001handbook}.

\paragraph{More complex constraints.}
As just discussed, constraints such as negation, logical implication and reverse implication already pose challenges to state-of-the-art LLMs in terms of consistency.
While we will focus on the Llama 2 LLM family in this work, similar shortcomings have been highlighted  for even larger models such as ChatGPT and GPT-4 \citep{jang2023consistency}.
Nevertheless, they constitute only a small fraction of the possible real-world reasoning scenarios LLMs can be asked to deal with.
Consider for example the following textual representations of facts, as extracted from EntailmentBank~\citep{dalvi2022explaining}:
$f_1:$ ``melting is a kind of phase change'',
$f_2:$ ``the ice melts'',
$f_3:$ ``the ice undergoes a phase change'',
$f_4:$ ``phase changes do not change mass'',
$f_5:$ ``the mass of the ice will not change''.
They
obey the following logical constraint
\begin{equation}
    \label{eq:entailment-bank}
    (z_{f_1} \land z_{f_2} \rightarrow z_{f_3}) \land z_{f_4} \rightarrow z_{f_5}.
\end{equation}
In the next section, we will introduce our general framework that can improve logical consistency of fine-tunable LLMs w.r.t.
\textit{any} logical constraint expressible in propositional logic.

\section{Logically-consistent LLMs via NeSy integration}
\label{sec:methodology}

We assume we are given a KB comprising a limited set of textual statements and associated truth values $\mathcal{D}_F = \{(f_1, z_{f_1}^{*}) \ldots, (f_n, z_{f_n}^{*})\}$, encoding simple facts such as ``an albatross is a bird'' (true) and ``a computer is a bird'' (false),
and a set of logical constraints $\mathcal{D}_C = \{\alpha_1,\ldots,\alpha_m\}$ defined over facts in $\mathcal{D}_F$, comprising implications, negations or more complex constraints as defined in \cref{sec:background}.

Given a pre-trained LLM encoding a distribution $p_\theta$ over tokens, our objective is to fine-tune it
to be more consistent w.r.t. $\mathcal{D}_F$, $\mathcal{D}_C$ and itself.
As an important side benefit, we expect the fine-tuned LLM to generalize to -- and be consistent with -- the truth values of unseen facts $f_{n+1}, f_{n+2}, \ldots,$ that can be either logically inferred by applying the constraints in $\mathcal{D}_C$ to $\mathcal{D}_F$ (e.g., by applying modus ponens) or
that are semantically similar to facts in $\mathcal{D}_F$.
For example, since albatross and cockerel are birds, and since this is reflected by their semantic similarity as encoded by the LLM, we expect an LLM consistent with the constraint (``an albatross is a bird'' $\rightarrow$ ``an albatross can fly'') to correctly infer that ``a cockerel can fly'' too.

A principled probabilistic approach to do so is to encourage the LLM $p_\theta$ to allocate all probability mass to configurations of truth values that are consistent with the constraints $\alpha_i\in\mathcal{D}_C$, for instance by penalizing it proportionally to the probability it allocates to inconsistent truth values for all facts in the KB.
For every $\alpha_i$, the total probability allocated to the consistent configurations is
\begin{equation}
    \label{eq:prob-constraint}
    \mathsf{Pr}(\alpha_i):=\mathbb{E}_{\mathbf{z}\sim p_\theta(\mathbf{z})}[\mathds{1}\{{\bf z} \models \alpha_i\}] = \sum\nolimits_{{\bf z} \models \alpha_i}\;\; p_\theta(\mathbf{z})
\end{equation}
where $\mathbf{z}$ is a vector containing the truth assignments $z_1,\ldots, z_K$ of all the $K$
facts appearing in the constraint $\alpha_i$, and ${\bf z} \models \alpha_i$ indicates
that the assignment ${\bf z}$ satisfies the constraint.
For example, consider two facts $f_1:$ ``a daffodil is a flower'' and  $f_2:$ ``a daffodil is mortal'' and the constraint $\alpha^{\prime}:z_{f_1} \rightarrow z_{f_2}$ stating that being a flower entails that the daffodil is mortal. 
Then, all the configurations of $\mathbf{z}=(z_{f_1}, z_{f_2})$ would satisfy $\alpha^{\prime}$ with the exception of $(\top, \bot)$ which clearly violates it. 
\cref{eq:prob-constraint} is a special instantiation of computing the weighted model count (WMC) \citep{chavira2008probabilistic,van2024independence} of a logical formula $\alpha_i$, where the weights associated to each model (a satisfying assignment to the formula) are given by the probabilities encoded by the LLM.

Furthermore, we can rewrite such probabilities $p_\theta(\mathbf{z})$ as the product the probabilities of the truth values of each fact, noting that for many LLM architectures they are conditionally independent given the embeddings at the last layer.
By taking the logarithm and reversing it into a minimization problem, we obtain the \textit{semantic loss} (SL) \citep{xu2018semantic} objective that our \methods minimize:
\begin{equation}
    \label{eq:semantic_loss}
    \tag{SL}
    \mathcal{L}(\alpha_i, p_{\theta}) = 
    - \log \sum\nolimits_{{\bf z} \models \alpha_i}\;\;\prod\nolimits_{j:{\bf z} \models z_{f_j}} p_\theta(z_{f_j})\;\;\prod\nolimits_{j:{\bf z} \models \lnot z_{f_j}} (1-p_\theta(z_{f_j}))
\end{equation}
where $j:{\bf z} \models z_{f_j}$ (resp. $j:{\bf z} \models \lnot z_{f_j}$) indicates that the $j$-th fact in $\alpha_i$ is associated $\top$ (resp. $\bot$).
Consider the implication constraint $\alpha^{\prime}$ as defined before for encoding that a daffodil is mortal for being a flower.
Its satisfying assignments are $\mathbf{z}\models\alpha^{\prime}\in\{(\top, \top), (\bot, \top), (\bot, \bot)\}$.
Then, the summation in \cref{eq:semantic_loss}
amounts to computing:
\begin{equation*}
    \label{applied_semantic_loss_1}
            p_\theta(z_{f_1} = \top) p_\theta(z_{f_2} = \top) +
                (1 - p_\theta(z_{f_1} = \top)) p_\theta(z_{f_2} = \top) +
                (1 - p_\theta(z_{f_1} = \top)) (1 - p_\theta(z_{f_2} = \top)))
\end{equation*}
where we can obtain the individual probabilities of facts being true directly by reading off the likelihood of utterances produced by the LLM, that is:
\begin{align*}
    p_\theta(z_{f_1} = \top )
        & = p_\theta(x_t = \ell_{\mathsf{true}} \mid x_1,\ldots,x_{t-1} = \text{``Is a daffodil a flower?''})
    \\
    p_\theta(z_{f_2} = \top )
        & = p_\theta(x_t = \ell_{\mathsf{true}} \mid x_1,\ldots,x_{t-1} = \text{``Is a daffodil a mortal?"}).
\end{align*}

In the case of a constraint such as \cref{eq:mp}, the inner summation of the SL would reduce to a single configuration $\mathbf{z}=(\top, \top)$ when $z_{f_1}^{*}=\top$, which 
can be interpreted as a special kind of cross-entropy computed only on pairs of facts considered to be jointly true in the KB, and to the set $\{(\bot, \top), (\bot, \bot)\}$ when $z_{f_1}^{*}=\bot$. 
Note that \cref{eq:semantic_loss} is \textit{agnostic to the kind of logical constraint involved}, and therefore makes our approach general enough to tackle several settings where consistency-preserving solutions have been devised for specific constraints \citep{burns2022discovering,kassner2023language,mitchell2022enhancing}.

Crucially, the procedure to compute the models of a logical constraint can be automated.
However, naively computing the sum in \cref{eq:semantic_loss} would require exponential time w.r.t. the number of possible facts in $\mathbf{z}$.
In fact, computing the WMC of a logical formula is a \#P-hard problem in general \citep{chavira2008probabilistic}.
However, thanks to recent advancements in neuro-symbolic reasoning, we can compute that probability and differentiate through it efficiently \citep{darwiche2011sdd, xu2018semantic, ahmed2022semantic}.
Specifically, we rely on modern \textit{compilers} that translate a logical formula $\alpha_i$ into compact and differential computational graphs called circuits \citep{darwiche2003differential, vergari2019tractable}, such as sentential decision diagrams \citep{darwiche2011sdd, oztok2015top, choi2013dynamic}.

To recap, during training we
loop over every constraint in $\alpha_i\in\mathcal{D}_{C}$, prompt the LLM to gather the probabilities of every fact participating in $\alpha_i$ to be true and plug them in our only loss, as described in \cref{eq:semantic_loss}.
Then, we
backpropagate as to fine-tune (some of) the parameters $\theta$ of the LLM, by using LoRA \citep{hu2021lora} and quantization \citep{dettmers2023qlora}  if necessary.
This simple and principled recipe is able to scale well and
is extremely effective
at improving logical consistency on a number of well-known benchmarks, as discussed in \cref{sec:exps}.

\section{Related Work}
\label{sec:related}

\paragraph{LLMs and factual reasoning.}
LLMs are
increasingly being employed as implict KBs \citep{petroni2019language,alkhamissi2022review},
however
ensuring they are factually consistent is still an open challenge \citep{wang2023survey,augenstein2023factuality}.
A number of works augment LLMs with external KBs, especially in the context of QA, and with the primary aim of improving answer factuality \citep{kassner2023language,mitchell2022enhancing,li2024flexkbqa}.  A popular approach to do so is retrieval augmented generation \citep{lewis2020retrieval,li2024enhancing}, which however is not yet suited for more complex reasoning scenarios.
Alternatively, external KBs have been used to improve reasoning, e.g., via prompt learning \citep{palagin2023ontochatgpt} or ex-post model editing \citep{shi2023mededit}. However, current knowledge editing methods, including supervised fine-tuning, do not guarantee the propagation of factuality between units of knowledge related by logical relations \cite{cohen2023evaluating, akyürek2024deductive}.
Mitigating hallucinations in LLMs \citep{andriopoulos2023augmenting,rawte2023survey} is related to enforcing factuality, but as generated inconsistencies might not map to a single entry in a KB, they are harder to detect and prevent \citep{hong2024hallucinations}.

\paragraph{More complex reasoning with LLMs.}
Much less attention has been posed to other forms of reasoning, such as combining modus ponens, consistent negation and combination thereof. 
Even when this is done, reasoning is generally cast as a QA task, where an LLM has to predict the satisfiability of logical formulas of different complexities.
To this end, benchmarks such as SimpleLogic \citep{zhang2022paradox}
 or LogicBench \citep{parmar2023logicbench}
have been proposed.
Implication or entailment \citep{maccartney2009natural,evans2018can} are also usually cast as a QA prediction task \citep{raj2023semantic}.
Datasets such as BeliefBank \citep{kassner2021beliefbank} provide collections of simple implication constraints to test this, while more sophisticated benchmarks such as EntailmentBank \citep{dalvi2022explaining} collect more complex implications, e.g., trees of natural language statements.
Shortcomings in consistent reasoning have been recently highlighted for larger LLMs such as ChatGPT and GPT-4 variants \citep{jang2023consistency}, which are however harder to fine-tune efficiently.
Other works \citep{berglund2023reversal} highlighted how (even large) LLMs suffer from not being able to recognize the logical equivalence of ``A is-a B'' relationships and ``B is-a A'' ones. These relationships could be seen as a type of logical constraint, specifically concept membership to an ontology class, and hence could be modeled in our framework.

For complex reasoning scenarios, logical consistency can be improved in a number of ways, the most successful of which involves external tools, such as MaxSAT solvers, which flip the predictions of an LLM to be (approximately) consistent with a set of related questions, as done by ConCoRD \citep{mitchell2022enhancing}.
Analogously, self-consistency can be ameliorated by first constructing a belief graph -- a factor graph relating the beliefs of an LLM fine-tuned on implications such as Entailer \citep{tafjord2022entailer} -- over which a MaxSAT solver is applied
\citep{kassner2023language}.
Higher level constraints can also be checked and enforced with external verifiers \citep{wang2024instructions}.
Differently from \methods, backpropagating through these external tools is hard \citep{poganvcic2019differentiation},
furthermore, while they can guarantee self-consistency among facts \textit{within} every call of a MaxSAT solver, this cannot be done for the same facts \textit{across} different calls.

\paragraph{Semantic loss \& other NeSy approaches}
There is a vast literature on NeSy integration methods \citep{de2019neuro,de2021statistical}, most of which are used for enforcing constraint on tabular data \citep{giunchiglia2020coherent}, image data \citep{xu2018semantic,shindo2021neuro,ahmed2022semantic} and more recently video recognition \citep{giunchiglia2023road} with the purpose of building trustworthy predictors.
Several variants of the semantic loss \citep{xu2018semantic,ahmed2022neuro,ahmed2024pseudo} and neural weighted model counting \citep{van2024independence} have been proposed but, to the best of our knowledge, never employed to enforce logical consistency in LLMs.
In our experiments we found that our simple formulation (\cref{eq:semantic_loss}) is good enough to greatly improve consistency over previous state-of-the-art methods in NLP (\cref{sec:exps}). 
Closer to our work, \citep{zhang2023improved} applied a semantic loss to instill first-order rule constraints in the embedding space of entities in encoder-only models to reason on the CLUTTR benchmark \citep{sinha2019clutrr}, comprising semi-synthetic stories involving hypothetical families.
Fuzzy logic approaches \citep{van2022analyzing} can be used to distill regularizers that can promote consistency \citep{li2019logicdriven}.
Differently from our probabilistic logic approach however, they are syntax-dependent, i.e., rewriting a constraint into a logically equivalent one  would yield a different penalty term and can greatly influence optimization \citep{van2022analyzing,di2020efficient}.

\section{Experiments}
\label{sec:exps}

We aim to answer the following research questions:
\textbf{RQ1}: Can \methods achieve comparable or superior consistency to methods using external reasoners using less training data?
\textbf{RQ2}: Can \methods retain good consistency to unseen types of constraint at training time? How much does training on all the constraints jointly improve consistency overall?
\textbf{RQ3}: Can \methods transfer consistent knowledge to domains out of the training distribution?

\subsection{RQ1: How do \methods perform compared to external solvers?}
\label{exp:rq1}

We reproduce the experimental setting of   Mitchell et al. \citep{mitchell2022enhancing} to compare against ConCoRD, a symbolic layer that uses a MaxSAT solver to impose self-consistency for implication ex-post.

\textbf{Data.} 
We train \methods on the BeliefBank \citep{kassner2021beliefbank}.
We use the three splits as in Mitchell et al. \citep{mitchell2022enhancing}:
a ``calibration"  set of $1,072$ annotated facts about $7$ entities of the form \textit{(subject, property, true/false)} 
used for training,
a ``silver" set of $12,636$ facts about 85 entities used for evaluation,
and a set of $2224$ valid abstract logical implications.
We generate ground implication rules ($\mathcal{D}_C$) by looking up the subjects of all facts in the training set: if the antecedent or the consequent fact of the general constraint is known for that subject, we add the subject ground implication constraint to the dataset.
\cref{app:data-prep-rq1} details the whole process.

To measure generalization across entities, we generate two controlled splits of the training calibration set:
\textit{T1 facts}, appearing either as antecedents or consequents in the constraints; \textit{T2 facts}, appearing exclusively as consequents.  
The goal is to correctly guess the consequents by seeing only the antecedents and the constraints. 
We subsequently test the effects of pure supervised fine-tuning on a portion of random facts from the whole calibration set (T1+T2).

\textbf{Models.} 
As in Mitchell et al. \cite{mitchell2022enhancing},
we use Macaw-Large \citep{tafjord2021generalpurpose} ($770M$ parameters), a sequence-to-sequence language model capable of multi-angle QA with fixed prompt templates.
We keep the same prompts used for Macaw, reported in Appendix \ref{prompts:macaw}.
At test time, we verify the validity of the answer format and consider any invalid or negative response as a belief with label "false".
We adopt a similar set of hyperparameters as for Macaw \cite{tafjord2021generalpurpose}: we fine-tune our models for $3$ epochs with a learning rate fixed to $\gamma = 3\cdot10^{-4}$, batch size $4$ with gradient accumulation ($64/16$ steps), on one nVidia A30 24GB GPU. We use AdamW \citep{loshchilov2016sgdr} as optimizer with a default weight decay $\lambda = 10^{-2}$.

\textbf{Competitors and Metrics.} We compare 
{ConCoRD} as applied to Macaw-Large, using RoBERTa-ANLI \citep{liu2019roberta} for relationship inference, versus 
a pre-trained {Macaw-Large} model from \cite{tafjord2021generalpurpose} as zero-shot baseline and our LoCo version of it (LoCo-Macaw).
We evaluate our models for \textit{factuality} and \textit{implication self-consistency}.
We measure the former with the $F_1$ score to account for the
unbalance between false and true facts \citep{kassner2021beliefbank}.
Factuality is measured on the two splits (antecedents and consequents) and the complete facts set (Tot) for both calibration and silver splits.
For \textbf{\textit{implication self-consistency}}, sometimes named just ``consistency'' \cite{li2019logicdriven}, we query beliefs from LLMs about the complete facts set and count the fraction of violated constraints in $\mathcal{D}_C^{\text{test}}$ according to the implication rule (\ref{eq:imp}), that is, when a true antecedent for the model implies a false consequent, to then compute:
\begin{equation}
    \label{eq:log-consistency}
    1 - | \{ \alpha_i = (z_j \rightarrow z_k) : z_j = \top, z_k = \bot \}|
    \ / \
    | \{ \alpha_i = (z_j \rightarrow z_k) : z_j = \top \}|.
\end{equation}
\textbf{Results.} 
\cref{tab:RQ1:test} reports all metrics for all models. 
We firstly observe a net improvement in both factuality and logical consistency with our \methods, compared to pre-trained Macaw-Large and the ConCoRD variant.
Standard supervised fine-tuning with the XENT loss on antecedent facts is insufficient: due to a class imbalance between true facts ($\sim{10\%}$) and false facts ($\sim{90\%}$), the model tends to label any statement as ``false''.
This is accentuated in the training distribution (see \cref{tab:RQ1:train}).
Assuming the language model can access to a portion of consequent facts, \methods still yields better logical consistency and factuality for unseen consequents in low-data regimes (e.g., 5-10\% of the T1+T2 dataset) compared to canonical supervised fine-tuning. 
When they are allowed to see more data (e.g., 75\% of the T1+T2 dataset), traditionally fine-tuned models can ``cheat'' and directly learn about the consequents (somehow equivalent to memorizing a single row of the truth table).
In this scenario, \methods achieve comparable logical self-consistency and factuality over consequents, but less on the antecedents.

\begin{table}[!t]
    \begin{minipage}{0.32\textwidth}
        \caption{
             \textbf{\methods achieve better logical self-consistency and factuality than ConCoRD \citep{mitchell2022enhancing}} as measured via \cref{eq:log-consistency} and $F_1$ scores when fine-tuned only on T1 facts only and boost performance in the presence of a small fraction of T1+T2 facts (5-10\%).
            A similar trend is visible on training data (\cref{tab:RQ1:train}).
         }
         \label{tab:RQ1:test}
    \end{minipage}\hfill\begin{minipage}{0.66\textwidth}
        \scriptsize
        \sc
        \begin{tabular}{cccccc}
            \toprule 
                    Method & Train Subset & Ant $F_1$ & Con $F_1$ & Tot $F_1$ &  Imp \\
            \midrule
                    \multicolumn{1}{l}{ConCoRD} & ~ & ~ & ~ & 0.91 & 0.91 \\
                    \multicolumn{1}{l}{Macaw-Large} & ~ & 0.52 & 0.90 & 0.81 & 0.83 \\ 
                    \multicolumn{1}{l}{Macaw+XENT} & T1 & 0.13 & 0.01 & 0.03 & 0.72 \\
                    \multicolumn{1}{l}{LoCo-Macaw} & T1 & \textbf{0.79} & \textbf{0.98} & \textbf{0.96} & \textbf{0.99} \\
            \midrule
                    \multicolumn{1}{l}{Macaw+XENT} & T1+T2 (5\%) & 0.23 & 0.78 & 0.72 & 0.82 \\ 
                    \multicolumn{1}{l}{LoCo-Macaw} & T1+T2 (5\%) & \textbf{0.67} & \textbf{0.83} & \textbf{0.81} & \textbf{0.92} \\
            \midrule
                    \multicolumn{1}{l}{Macaw+XENT} & T1+T2 (10\%) & \textbf{0.55} & \textbf{0.97} & \textbf{0.91} & 0.90 \\
                    \multicolumn{1}{l}{LoCo-Macaw} & T1+T2 (10\%) & 0.45 & \textbf{0.97} & 0.89 & \textbf{0.93} \\
            \midrule
                    \multicolumn{1}{l}{Macaw+XENT} & T1+T2 (75\%) & \textbf{0.85} & \textbf{0.99} & \textbf{0.97} & \textbf{0.98} \\
                    \multicolumn{1}{l}{LoCo-Macaw} & T1+T2 (75\%) & 0.79 & \textbf{0.99} & 0.95 & \textbf{0.98} \\
            \bottomrule
        \end{tabular}
    \end{minipage}
        \begin{center}\sc
        \end{center}
\end{table}

In conclusion, we observe our fine-tuning method allows Macaw-large to be more logically self-consistent than with an external solver.
We conjecture that this is possible thanks to the high semantic similarity between facts in the train and test splits (\cref{sem-op:bb-bb}).
In terms of inference speed, our \methods take less time that querying the same base model and an additional reasoner\footnote{On BeliefBank, \methods take 2405.28s at test time, compared to ConCoRD \cite{mitchell2022enhancing}, 3669.33s.}, at the cost of a one-time training step that can be amortized.\footnote{Training \method takes 2124.48s on BeliefBank.}
Moreover, our semantic loss is more sample-efficient than XENT fine-tuning to achieve higher logical consistency especially with small portions of ground-truth data.

\subsection{RQ2: How do \methods deal with different logical constraints?}
\label{exp:rq2}

\textbf{Setting.} As in \cref{exp:rq1}, we use BeliefBank to train and evaluate our \methods on different types of logical rules. We use $90\%$ and $10\%$ of \textit{T1 facts} for training and validation, respectively; $\textit{T2 facts}$ for testing. We employ two sets of labels to make our models less sensitive to the prompt format; at training time, one format is chosen with $50\%$ chance for each batch; details in Appendix \ref{prompts:loco-lms}. At test time we do not apply any strict parsing on the outputs: unless the token encodes the truth label (e.g., ``Is a computer an electronic device? \textbf{yes}''), the output is considered as a negative answer.

\textbf{Models.} To train larger language models, we choose the LLaMa-2 \citep{touvron2023llama} family of decoder-only models, widely adopted in literature for its performance across a variety of tasks and domains. 
We consider three baselines: the available pre-trained $7$b and $70$b models, 4-bit NormalFloat quantized \cite{dettmers2023qlora}, with greedy sampling strategy, temperature $t = 1.0$ and dropout disabled; we also perform supervised fine-tuning of the $7$b model (4-bit, with LoRA \citep{hu2021lora}) on the ground truth T1+T2 facts set, namely ``LLaMa-2-7b + XENT''. 
We derive our \methods fine-tuning with our proposed method LLaMa-2 7b, with 4-bit quantization and LoRA. 
We limit the generation to $4$ tokens following the input. 
We adopt a similar set of hyperparameters to LoRA: we fine-tune our models for $5$ epochs keeping the learning rate fixed to $\gamma = 3\cdot10^{-4}$, batch size $64$, on $1$ nVidia A100-40GB GPU. We use AdamW \citep{loshchilov2016sgdr} as optimizer with a default weight decay $\lambda = 10^{-2}$.
We use the SL to finetune three \method variants: for negation (\ref{eq:neg}), factual implication consistency (\ref{eq:mp}) and their conjunction, i.e., given an implication $f_1\rightarrow f_2$ we provide the SL with the constraint:
\begin{equation}
    \label{eq:super}
    \tag{\textsc{Super}}
     (z_{f_1}\xor z_{\widetilde{f_1}}) \wedge (z_{f_1}=z_{f_1}^{*})\wedge (z_{f_1} \rightarrow z_{f_2}) \wedge
     (z_{f_2}\xor z_{\widetilde{f_2}})
\end{equation}
where $\widetilde{f_1}$ and $\widetilde{f_2}$ encode the textual negation of $f_1$ and $f_2$, generated via ConCoRD's templates.

\textbf{Metrics.} We fine-tune on \ref{eq:neg}, \ref{eq:mp} or \ref{eq:super} and evaluate on all constraints.
Specifically, we measure the implication self-consistency, defined in \cref{eq:log-consistency}, as well as the \textbf{\textit{implication consistency}}:
\begin{equation}
    \label{eq:imp-consistency}
    1 - | \{ \alpha_i = (z_j \rightarrow z_k) : z_j^* = \top, z_k = \bot \}|
    \ / \
    | \{ \alpha_i = (z_j \rightarrow z_k) : z_j^* = \top \}|
\end{equation}
where $z_j^*$ is the ground truth value of a fact.
We also measure \textbf{\textit{reverse implication consistency}}
\begin{equation}
    \label{eq:rev-consistency}
    1 - | \{ \alpha_i = (z_{\widetilde{k}} \rightarrow z_{\widetilde{j}}) : \neg z_k^* = \top, z_{\widetilde{j}} = \top \}|
    \ / \
    | \{ \alpha_i = (z_{\widetilde{k}} \rightarrow z_{\widetilde{j}}) : \neg z_k^* = \top \}|
\end{equation}
and the \textbf{\textit{reverse implication self-consistency}} variant:
\begin{equation}
    \label{eq:self-rev-consistency}
    1 - | \{ \alpha_i = (z_{\widetilde{k}} \rightarrow z_{\widetilde{j}}) : z_{\widetilde{k}}= \bot, z_{\widetilde{j}} = \top \}|
    \ / \
    | \{ \alpha_i = (z_{\widetilde{k}} \rightarrow z_{\widetilde{j}}) : z_{\widetilde{k}} = \bot \}|
\end{equation}
where $z_{\widetilde{k}}$ and $z_{\widetilde{j}}$ are the truth values of the textual negations of facts $k$ and $j$ according to the model.
For negation self-consistency we compute
        \begin{equation}
        \label{eq:log-negation-consistency}
            1 - { |\{ \alpha_i = (z_j \xor z_{\widetilde{j}}): z_j = z_{\widetilde{j}}}\}|   \ /\  {|\alpha_i = (z_j \xor {z_{\widetilde{j}}})|}.
        \end{equation}
As in \cref{exp:rq1}, we measure factuality (FAC) as the $F_1$ score on a set of ground truth facts. 
Finally, we account for possible shifts in the language modeling distribution by computing its perplexity (PPL) on WikiText \citep{merity2016pointer}, formatted as a single token sequence.

\textbf{Results.} In \cref{tab:RQ2:avg_test}, we first observe an overall boost in factuality for all \methods over the 7b baselines. 
Compatibly with \cref{tab:RQ1:test}, supervised fine-tuning is not sufficient to improve logical consistency significantly. 
Our \method trained exclusively on \ref{eq:imp} constraints  performs best in factuality and implication consistency; however, as we haven't trained it on negated facts, scores on negation consistency and reverse implication are notably low.
Finally, fine-tuning a \method on the combination of both constraints (\ref{eq:super}), yields on average the most consistent language model, which on average surpasses even Llama 2 70B, a much larger model.
Overall, fine-tuning with our method does not impact negatively fluency, as  measured by perplexity.

\begin{table}[!t]
	\centering
    
        \scriptsize
        \caption{
            \textbf{\methods achieve higher (self-)consistency than off-the-shelf baselines and models trained with supervised fine-tuning} (+XENT) on the BeliefBank test split. 
            Scores are averaged across two sets of prompts and truth labels, for which results are reported in Appendix \ref{table_0_silver} and \ref{table_1_silver}. 
        }
        \label{tab:RQ2:avg_test}
        \begin{center}
        \setlength{\tabcolsep}{3pt}
        \sc
            \begin{tabular}{lcrrrrrrrrr}
                \toprule 
                        {} & {} & {} & \multicolumn{3}{c}{consistency} & \multicolumn{3}{c}{self-consistency} & {} \\
                        \cmidrule(lr){4-6}
                        \cmidrule(lr){7-9}
                        model & train & PPL & FAC & IMP & REV & NEG & IMP & REV & AVG \\
                \midrule
                        LLaMa-2-7b Zero Shot & ~ & 62.41 & 0.39 & 0.52 & 0.13 & 0.42 & 0.30 & 0.15 & 0.32 \\
                        LLaMa-2-7b Few Shot & ~ & 52.30 & 0.53 & 0.71 & 0.34 & 0.38 & 0.48 & 0.47 & 0.48 \\
                        LLaMa-2-7b CoT & ~ & 52.30 & 0.52 & 0.64 & 0.67 & 0.40 & 0.64 & 0.67 & 0.59 \\
                        LLaMa-2-70b Zero Shot & ~ & 44.90 & 0.47 & 0.69 & 0.81 & 0.13 & 0.31 & 0.91 & 0.55 \\
                \midrule       
                        LLaMa-2-7b + XENT & T1+T2 & 116.85 & 0.25 & 0.46 & 0.01 & 0.07 & 0.81 & 0.01 & 0.27 \\
                        LoCo-LLaMa-2-7b (NEG) & T1 & 62.21 & 0.44 & 0.65 & 0.43 & \textbf{0.96} & 0.28 & 0.36 & 0.52 \\ 
                        LoCo-LLaMa-2-7b (F-IMP) & T1 & 67.15 & \textbf{0.99} & \textbf{0.99} & 0.07 & 0.00 & \textbf{0.99} & 0.07 & 0.51 \\ 
                        LoCo-LLaMa-2-7b (Super) & T1 & 62.23 & 0.74 & 0.77 & \textbf{0.77} & 0.87 & 0.71 & \textbf{0.77} & \textbf{0.77} \\ 
                \bottomrule
            \end{tabular}
        \end{center}
\end{table}

\subsection{RQ3: Can finetuning \methods help consistency on unseen KB?}
\label{exp:rq3}

\textbf{Data.} We evaluate \methods on the EntailmentBank \citep{dalvi2022explaining} test split, as proposed by Kassner et al. \cite{kassner2023language} to reason on graphs of logical entailments. It consists of $302$ implication trees spawning 805 constraints, with an average of $6.57$ statement nodes and $2.66$ constraints per tree; we consider each node of each tree as a statement with natural language with truth label set to $1$. We limit the tree depth to $5$. An illustrated example is provided in Appendix \ref{viz:eb_example0}. As in \ref{exp:rq2}, we test two prompt and label formats. We assume that a possible semantic overlap between the training and test distributions, BeliefBank and EntailmentBank respectively, could underlie higher consistency scores across entailment trees; we estimate such overlap in Appendix \ref{sem-op:bb-eb}.

\textbf{Competitors and Metrics.} 
We test our \methods based on LLaMa-2 7b and previously trained in \ref{exp:rq2} on BeliefBank, without applying any changes. As baseline model, we consider LLaMa-2 7b without quantization.
This experimental setup is inspired by Kassner et al. \cite{kassner2023language}, from whom we derive the notion of self-consistency on trees of entailments: each entailment tree $t \in \mathcal{T}$ is a direct acyclic graph with a single root encoding the hypothesis to be proved;  a subtree $t'$ consists in each parents-child relationship in $t$, representing an entailment between the parent nodes (antecedents in logical conjunction) and the child (consequent).
See \cref{viz:eb_example0} for an example.
For each tree $t$, we count the amount of violated subtrees $t'$, that is when a true conjunction of antecedents does not imply a true consequent. Finally, we measure logical consistency as the fraction of the total violated subtrees over the total number of subtrees in $\mathcal{T}$. 

\textbf{Results.} In Table \ref{tab:RQ3:eb} we report logical consistency across depths. Scores are averaged across two sets of prompts and labels, detailed results are reported in Appendix \ref{app:RQ3:eb_truefalse}. We observe the consistency decreases across depths for the baseline model, until it flattens out, as more implications are evaluated. Conversely, \method (F-IMP) and \method (Super) achieve higher consistency across depths.
While promising, these results should be interpreted with caution, for two reasons. Firstly, we observed variability in model predictions with varying prompt formats and labels (Appendix \ref{app-tab:RQ3}), suggesting further engineering for more consistent answers. 
Second, while be measure a discrete semantic similarity between the two datasets (\cref{sem-op:bb-eb}) which can justify transfer, we note that our measure are cosine similarities and their effectiveness  might depend on the pre-training task \cite{Steck_2024}.
This encourages further research on employing neuro-symbolic methods to improve multi-hop consistency in LMs w.r.t. external KB \cite{kassner2023language} or the model's own implications \cite{akyürek2024deductive}.

\begin{table}[!t]
        \label{tab:RQ3:eb}
	\begin{minipage}{.5\textwidth}
	     \scriptsize
      \setlength{\tabcolsep}{3pt}
              \sc
            \begin{tabular}{lrrrrr}
                \toprule 
                         {} & \multicolumn{5}{c}{depth} \\
                         \cmidrule(lr){2-6} 
                         model & 1 & 2 & 3 & 4 & 5 \\
                \midrule
                         LLaMa-2-7b & 0.87 & 0.76 & 0.59 & 0.61 & 0.63 \\
                \midrule
                         LoCo-LLaMa-2-7b (NEG) & 0.51 & 0.51 & 0.51 & 0.52 & 0.52 \\
                         LoCo-LLaMa-2-7b (F-IMP) & \textbf{0.98} & \textbf{0.98} & \textbf{0.98} & \textbf{0.98} & \textbf{0.98} \\
                         LoCo-LLaMa-2-7b (Super) & 0.69 & 0.68 & 0.68 & 0.68 & 0.69 \\
                \bottomrule
            \end{tabular}
	\end{minipage}\hfill\begin{minipage}{.48\textwidth}
	            \caption{
            \textbf{\methods can be consistent across unseen trees of entailments} from EntailmentBank when trained for implication consistency (F-IMP) on BeliefBank.
            Finetuning for negation alone (NEG) does not seem to improve over the baseline.
        }
	\end{minipage}
\end{table}

\section{Discussion and Further Work}
\label{sec:discussion}

\paragraph{Limitations.}
One limitation of our approach is sensitivity to the choice of prompt format, a general phenomenon \citep{white2023prompt} that in our case means (self-)consistency improvements do not always carry over across formats.  This can be partially addressed by fine-tuning using a mixture of formats, as we do in \cref{sec:exps}.
While our \ref{eq:semantic_loss} is constraint-agnostic, in practice we
 fine-tune \methods only against a combination of implications and exclusive ORs.  
While this setup is already richer than those studied in related works (\cref{sec:related})
and achieves positive transfer to tasks requiring multiple reasoning steps, it leaves more room for future work on more complex benchmarks.

\methods fine-tuning relies on two assumptions: that the probabilities of facts are conditionally independent given the LLM inner state, and that the constraints in the KB are correct.  
The former readily applies to many LLMs,
but assuming independence can bias the solutions learned by the \ref{eq:semantic_loss} \citep{van2024independence}.
For the latter, most KBs are well-curated, but fine-tuning models against incorrect or inconsistent rules can compromise consistency and fluency.  
Naturally, malicious users could intentionally train \methods against invalid rules to steer the model towards logical conclusions of their choice or potential reasoning shortcuts \citep{marconato2024not,marconato2024bears,bortolotti2024benchmark}. 

Our results show that \methods have improved (self-)consistency compared to recently introduced consistency layers which rely on external solvers, such as ConCoRD.
In future work, we plan to extend our analysis to more complex logical operators \citep{vergari2021compositional} and to consider more advanced probabilistic reasoning techniques that sport improved consistency guarantees \citep{ahmed2022semantic}.  
Another promising direction we have not explored is that of 
first materializing the beliefs of an LLM such as in REFLEX \citep{kassner2023language} and variants \citep{akyürek2024deductive} and use the \ref{eq:semantic_loss} to improve consistency while potentially storing and re-using derived rules in a writable external KB \citep{modarressi2023ret,modarressi2024memllm}.

\begin{ack}
Funded by the European Union. Views and opinions expressed are however those of the author(s) only and do not necessarily reflect those of the European Union or the European Health and Digital Executive Agency (HaDEA). Neither the European Union nor the granting authority can be held responsible for them. Grant Agreement no. 101120763 - TANGO.
AV is supported by the "UNREAL: Unified Reasoning Layer for Trustworthy ML" project (EP/Y023838/1) selected by the ERC and funded by UKRI EPSRC.
\end{ack}

\bibliography{refs}
\bibliographystyle{plain}

\newpage

\appendix

\section{Detailed setting and results}
\label{app:detailed-res}

\subsection{RQ1}
\label{app-tab:RQ1}

\subsubsection{Data preprocessing}
\label{app:data-prep-rq1}
We train \methods on the BeliefBank \citep{kassner2021beliefbank}, calibration split. 
This dataset is derived from ConceptNet \citep{speer2018conceptnet}, a large curated knowledge graph encoding factual knowledge and logical relations between entities at different levels of abstraction; we use the splits introduced by Mitchell et al. \citep{mitchell2022enhancing} for direct comparison. 
It consists of three pieces:
a ``calibration"  set of $1,072$ annotated facts about $7$ entities of the form \textit{(subject, property, true/false)} used for training,
a ``silver" set of $12,636$ facts about 85 entities used for evaluation,
and a set of $2224$ valid abstract logical implications.
To use our SL, we require defining a set of ground constraints.  
We derive these as follows.  For each general implication constraint, we lookup the subjects of all facts in the training set: if the antecedent or the consequent fact of the general constraint is known for that subject, we add the subject ground constraint to the dataset $\mathcal{D}_C$. 

We generate two splits: \textit{T1 facts}, appearing either as antecedents or consequents in the constraints; \textit{T2 facts}, appearing exclusively as consequents.  The goal is to correctly guess the consequents by seeing only the antecedents and the constraints. In the calibration set, we count $796$ antecedents and $276$ consequents, spawning $14,005$ grounded constraints. In the silver set, we count $9,504$ antecedents and $3,132$ consequents, spawning $169,913$ grounded constraints.
We subsequently test the effects of pure supervised fine-tuning: a portion of random facts from the calibration set (T1+T2) is taken with the goal to predict the excluded antecedent or consequent facts. 
We train on T1 facts and evaluate on T2 facts for RQ2 as well: \textit{T1 facts} (antecedents) constitute a valid subset for all the considered logical rules.

\begin{table}[!h]
        \label{tab:RQ1:train}
	\centering
	\renewcommand{\arraystretch}{1.2}
	\scriptsize
         \caption{
         \textbf{\methods achieve better logical self-consistency and factuality} as measured via \cref{eq:log-consistency} and $F_1$ scores when compared to cross-entropy fine-tuning (XENT) and baselines using external reasoners such as ConCoRD \citep{mitchell2022enhancing}  measured on train (calibration set) facts.
         For RQ1 (\cref{sec:exps}), 
         \methods fine-tuned on T1 facts only outperform training-free baseline for all metrics. 
         For RQ2, they boost performance in the presence of a small fraction of T1+T2 facts (5-10\%).
         For larger dataset sizes, \methods are competitive for consistency and factuality on consequents.
         }
        \begin{center}
            \begin{tabular}{ccccccc}
                \toprule 
                        {} & Method & Train size & Antecedents $F_1$ & Consequents $F_1$ & Total $F_1$ & Logical consistency \\
                \hline  
                        \parbox[t]{2mm}{\multirow{4}{*}{\rotatebox[origin=c]{0}{RQ1}}}
                        {} & \multicolumn{1}{l}{ConCoRD} & {} & {} & {} & 0.91 & 0.91 \\
                        {} & \multicolumn{1}{l}{MACAW} & {} & 0.47 & 0.84 & 0.78 & 0.82 \\
                        {} & \multicolumn{1}{l}{MACAW+XENT} & T1 & 0.46 & 0.08 & 0.14 & 0.79 \\ 
                        {} & \multicolumn{1}{l}{\method} & T1 & \textbf{0.98} & \textbf{0.99} & \textbf{0.99} & \textbf{1.00} \\ 
                \midrule
                        \parbox[t]{2mm}{\multirow{6}{*}{\rotatebox[origin=c]{0}{RQ2}}}
                        {} & \multicolumn{1}{l}{MACAW+XENT} & T1+T2 (5\%) & 0.31 & 0.73 & 0.69 & 0.90 \\ 
                        {} & \multicolumn{1}{l}{\method} & T1+T2 (5\%) & \textbf{0.34} & \textbf{0.77} & \textbf{0.72} & \textbf{0.92} \\ 
                \cmidrule(l){2-7}
                        {} & \multicolumn{1}{l}{MACAW+XENT} & T1+T2 (10\%) & 0.48 & 0.88 & 0.85 & 0.87 \\  
                        {} & \multicolumn{1}{l}{\method} & T1+T2 (10\%) & \textbf{0.52} & \textbf{0.95} & \textbf{0.89} & \textbf{0.91} \\ 
                \cmidrule(l){2-7}
                        {} & \multicolumn{1}{l}{MACAW+XENT} & T1+T2 (75\%) & \textbf{0.69} & \textbf{1.00} &\textbf{0.97} & 0.97 \\
                        {} & \multicolumn{1}{l}{\method} & T1+T2 (75\%) & 0.65 & \textbf{1.00} & \textbf{0.97} & \textbf{0.99} \\         
                \bottomrule
            \end{tabular}
        \end{center}
\end{table}

\newpage
\subsection{RQ2}
\label{app-tab:RQ2}

\begin{table}[!h]
	\centering
	\renewcommand{\arraystretch}{1.2}
        \scriptsize
        \caption{
            \textbf{\methods evaluated on BeliefBank, training (calibration) split.} Scores are averaged across two sets of prompts and truth labels. We observe fine-tuning with our method allows for higher logical consistency to different rules.
        }
        \label{table_0_calibratio_avg}
        \begin{center}
        \sc
            \begin{tabular}{lcrrrrrrrrr}
                \toprule 
                        {} & {} & {} & \multicolumn{3}{c}{consistency} & \multicolumn{3}{c}{self-consistency} & {} \\
                        \cmidrule(lr){4-6}
                        \cmidrule(lr){7-9}
                        model & train subset & PPL & FAC & IMP & REV & NEG & IMP & REV & AVG \\
                \midrule
                        LLaMa-2-7b Zero Shot & ~ & 62.41 & 0.41 & 0.57 & 0.21 & 0.42 & 0.28 & 0.24 & 0.36 \\
                        LLaMa-2-7b Few Shot & ~ & 52.30 & 0.52 & 0.70 & 0.45 & 0.38 & 0.48 & 0.46 & 0.50 \\
                        LLaMa-2-7b CoT & ~ & 52.30 & 0.52 & 0.64 & 0.67 & 0.40 & 0.64 & 0.67 & 0.59 \\   
                        LLaMa-2-70b Zero Shot & ~ & 44.90 & 0.49 & 0.72 & 0.80 & 0.12 & 0.32 & 0.91 & 0.56 \\
                \midrule        
                        LLaMa-2-7b + XENT & T1+T2 & 116.85 & 0.21 & 0.39 & 0.01	 & 0.10	 & 0.44	 & 0.01	 & 0.20 \\
                        LoCo-LLaMa-2-7b (NEG) & T1 & 62.21 & 0.28 & 0.52 & 0.43 & 0.82 & 0.55 & 0.36 & 0.49 \\
                        LoCo-LLaMa-2-7b (F-IMP) & T1 & 67.15 & \textbf{1.00} & \textbf{1.00} & 0.08 & 0.00 & \textbf{1.00} & 0.08 & 0.53 \\
                        LoCo-LLaMa-2-7b (Super) & T1 & 62.23 & 0.86 & 0.89 & \textbf{0.76} & \textbf{0.88} & 0.80 & \textbf{0.77} & \textbf{0.83} \\
                \bottomrule
            \end{tabular}
        \end{center}
\end{table}

\begin{table}[!h]
	\centering
	\renewcommand{\arraystretch}{1.2}
        \scriptsize
        \caption{
            \textbf{\methods evaluated on BeliefBank, training (calibration) split.} Prompt format 1 \texttt{[true, false]} is used. We observe fine-tuning with our method allows for higher logical consistency to different rules.
        }
        \label{table_0_calibration}
        \begin{center}
        \sc
            \begin{tabular}{lcrrrrrrrrr}
                \toprule 
                         {} & {} & {} & \multicolumn{3}{c}{consistency} & \multicolumn{3}{c}{self-consistency} & {} \\
                        \cmidrule(lr){4-6}
                        \cmidrule(lr){7-9}
                         model & train subset & PPL & FAC & IMP & REV & NEG & IMP & REV & AVG \\
                \midrule
                         LLaMa-2-7b Zero Shot & ~ & 62.41 & 0.43 & 0.63 & 0.33 & 0.38 & 0.29 & 0.39 & 0.41 \\
                         LLaMa-2-7b Few Shot & ~ & 52.30 & 0.53 & 0.74 & 0.36 & 0.28 & 0.42 & 0.37 & 0.45 \\
                         LLaMa-2-7b CoT & ~ & 52.30 & 0.67 & 0.76 & 0.77 & 0.32 & 0.74 & 0.77 & 0.66 \\ 
                         LLaMa-2-70b Zero Shot & ~ & 44.90 & 0.52 & 0,76 & 0.79 & 0.18 & 0.35 & 0.90 & 0.58 \\
                \midrule        
                         LLaMa-2-7b + XENT & T1+T2 & 116.85 & 0.37 & 0.47 & 0.02 & 0.16 & 0.89 & 0.02 & 0.32 \\
                         LoCo-LLaMa-2-7b (NEG) & T1 & 62.21 & 0.46 & 0.70	& \textbf{0.85} & 0.93 & 0.28 & 0.72 & 0.66 \\
                         LoCo-LLaMa-2-7b (F-IMP) & T1 & 67.15 & \textbf{1.00} & \textbf{1.00} & 0.08 & 0.00 & \textbf{1.00} & 0.08 & 0.53 \\
                         LoCo-LLaMa-2-7b (Super) & T1 & 62.23 & 0.88 & 0.91 & 0.72 & \textbf{0.94} & 0.86 & \textbf{0.73} & \textbf{0.84} \\
                \bottomrule
            \end{tabular}
        \end{center}
\end{table}

\begin{table}[!h]
	\centering
	\renewcommand{\arraystretch}{1.2}
        \scriptsize
        \caption{
            \textbf{\methods evaluated on BeliefBank, training (calibration) split.} Prompt format 2 \texttt{[yes, no]} is used. We observe fine-tuning with our method allows for higher logical consistency to different rules.
        }
        \label{table_1_calibration}
        \begin{center}
        \sc
            \begin{tabular}{lcrrrrrrrr}
                \toprule 
                         {} & {} & {} & \multicolumn{3}{c}{consistency} & \multicolumn{3}{c}{self-consistency} & {} \\
                        \cmidrule(lr){4-6}
                        \cmidrule(lr){7-9}
                         model & train subset & PPL & FAC & IMP & REV & NEG & IMP & REV & AVG\\
                \midrule
                         LLaMa-2-7b Zero Shot & ~ & 62.41 & 0.39 & 0.51 & 0.08 & 0.46 & 0.27 & 0.09 & 0.30 \\
                         LLaMa-2-7b Few Shot & ~ & 52.30 & 0.52 & 0.66 & 0.55 & 0.48 & 0.55 & 0.55 & 0.55 \\
                         LLaMa-2-7b CoT & ~ & 52.30 & 0.38 & 0.52 & 0.57 & 0.48 & 0.54 & 0.57 & 0.51 \\
                         LLaMa-2-70b Zero Shot & ~ & 44.90 & 0.46 & 0.68 & 0.81 & 0.05 & 0.28 & 0.93 & 0.54 \\
                \midrule
                        LLaMa-2-7b + XENT & T1+T2 & 116.85 & 0.05 & 0.32 & 0.00 & 0.04 & 0.00 & 0.00 & 0.07 \\
                        LoCo-LLaMa-2-7b (NEG) & T1 & 62.21 & 0.09 & 0.33 & 0.00 & 0.70 & 0.82 & 0.00 & 0.32 \\
                        LoCo-LLaMa-2-7b (F-IMP) & T1 & 67.15 & \textbf{1.00} & \textbf{1.00} & 0.08 & 0.00 & \textbf{1.00} & 0.08 & 0.53 \\
                        LoCo-LLaMa-2-7b (Super) & T1 & 62.23 & 0.84 & 0.87 & \textbf{0.79} & \textbf{0.82} & 0.74 & \textbf{0.80} & \textbf{0.81} \\
                \bottomrule
            \end{tabular}
        \end{center}
\end{table}

\begin{table}[!h]
	\centering
	\renewcommand{\arraystretch}{1.2}
        \scriptsize
        \caption{
            \textbf{\methods evaluated on BeliefBank, test (silver) split.} Prompt format 1 \texttt{[true, false]} is used. We observe fine-tuning with our method allows for higher logical consistency to different rules.
        }
        \label{table_0_silver}
        \begin{center}
        \sc
            \begin{tabular}{lcrrrrrrrr}
                \toprule 
                         {} & {} & {} & \multicolumn{3}{c}{consistency} & \multicolumn{3}{c}{self-consistency} & {} \\
                        \cmidrule(lr){4-6}
                        \cmidrule(lr){7-9}
                         model & train subset & PPL & FAC & IMP & REV & NEG & IMP & REV & AVG\\
                \midrule
                         LLaMa-2-7b Zero Shot & ~ & 62.41 & 0.41 & 0.55 & 0.22 & 0.41 & 0.30 & 0.25 & 0.36 \\
                         LLaMa-2-7b Few Shot & ~ & 52.30 & 0.53 & 0.75 & 0.37 & 0.27 & 0.41 & 0.37 & 0.45 \\
                         LLaMa-2-7b CoT & ~ & 52.30 & 0.67 & 0.76 & 0.77 & 0.32 & 0.74 & 0.77 & 0.67 \\
                         LLaMa-2-70b Zero Shot & ~ & 44.90 & 0.50 & 0.72 & 0.80 & 0.20 & 0.34 & 0.89 & 0.58 \\
                \midrule
                        LLaMa-2-7b + XENT & T1+T2 & 116.85 & 0.40 & 0.52 & 0.02 & 0.11 & 0.82 & 0.02 & 0.31 \\
                        LoCo-LLaMa-2-7b (NEG) & T1 & 62.21 & 0.44 & 0.64 & \textbf{0.86} & \textbf{0.92} & 0.28 & \textbf{0.72} & 0.64 \\
                        LoCo-LLaMa-2-7b (F-IMP) & T1 & 67.15 & \textbf{0.98} & \textbf{0.98} & 0.07 & 0.00 & \textbf{0.98} & 0.07 & 0.51 \\
                        LoCo-LLaMa-2-7b (Super) & T1 & 62.23 & 0.75 & 0.78 & 0.72 & 0.91 & 0.74 & \textbf{0.72} & \textbf{0.77} \\
                \bottomrule
            \end{tabular}
        \end{center}
\end{table}

\begin{table}[!h]
	\centering
	\renewcommand{\arraystretch}{1.2}
        \scriptsize
        \caption{
            \textbf{\methods evaluated on BeliefBank, test (silver) split.} Prompt format 2 \texttt{[yes, no]} is used. We observe fine-tuning with our method allows for higher logical consistency to different rules.
        }
        \label{table_1_silver}
        \begin{center}
        \sc
            \begin{tabular}{lcrrrrrrrr}
                \toprule 
                         {} & {} & {} & \multicolumn{3}{c}{consistency} & \multicolumn{3}{c}{self-consistency} & {} \\
                        \cmidrule(lr){4-6}
                        \cmidrule(lr){7-9}
                         model & train subset & PPL & FAC & IMP & REV & NEG & IMP & REV & AVG\\
                \midrule
                         LLaMa-2-7b Zero Shot & ~ & 62.41 & 0.37 & 0.48 & 0.04 & 0.43 & 0.29 & 0.04 & 0.28 \\
                         LLaMa-2-7b Few Shot & ~ & 52.30 & 0.53 & 0.67 & 0.57 & 0.49 & 0.58 & 0.53 & 0.56 \\
                         LLaMa-2-7b CoT & ~ & 52.30 & 0.38 & 0.52 & 0.57 & 0.48 & 0.54 & 0.57 & 0.51 \\
                         LLaMa-2-70b Zero Shot & ~ & 44.90 & 0.44 & 0.65 & 0.82 & 0.05 & 0.29 & 0.93 & 0.53 \\
                \midrule
                        LLaMa-2-7b + XENT & T1+T2 & 116.85 & 0.11 & 0.39 & 0.00 & 0.03 & 0.80 & 0.00 & 0.22 \\
                        LoCo-LLaMa-2-7b (NEG) & T1 & 62.21 & 0.44 & 0.65 & 0.00 & \textbf{1.00} & 0.28 & 0.00 & 0.40 \\
                        LoCo-LLaMa-2-7b (F-IMP) & T1 & 67.15 & \textbf{0.99} & \textbf{0.99} & 0.07 & 0.00 & \textbf{0.99} & 0.07 & 0.52 \\
                        LoCo-LLaMa-2-7b (Super) & T1 & 62.23 & 0.73 & 0.75 & \textbf{0.81} & 0.83 & 0.67 & \textbf{0.82} & \textbf{0.77} \\
                \bottomrule
            \end{tabular}
        \end{center}
\end{table}

\begin{table}[!h]
        \label{app:RQ3:eb_truefalse}
	\centering
	\renewcommand{\arraystretch}{1.2}
        \scriptsize
        \caption{
            \textbf{\methods can achieve higher consistency across depth than the baseline.} Scores are computed with Format 1 \texttt{[true, false]}, reported in Appendix \ref{prompts:loco-lms}. \method fine-tuned with on the implication rule achieves best consistency.
        }
        \begin{center}
        \sc
            \begin{tabular}{lrrrrr}
                \toprule 
                         {} & \multicolumn{5}{c}{depth} \\
                         \cmidrule(lr){2-6} 
                         model & 1 & 2 & 3 & 4 & 5 \\
                \midrule
                         LLaMa-2-7b & 0.73 & 0.77 & 0.79 & 0.80 & 0.80 \\
                \midrule
                         LoCo-LLaMa-2-7b (NEG) & 0.03 & 0.03 & 0.03 & 0.04 & 0.05 \\
                         LoCo-LLaMa-2-7b (F-IMP) & \textbf{0.97} & \textbf{0.96} & \textbf{0.97} & \textbf{0.97} & \textbf{0.97} \\
                         LoCo-LLaMa-2-7b (Super) & 0.75 & 0.74 & 0.73 & 0.73 & 0.74 \\
                \bottomrule
            \end{tabular}
        \end{center}
\end{table}

\newpage
\subsection{RQ3}
\label{app-tab:RQ3}

\begin{table}[!h]
        \label{app:RQ3:eb_yesno}
	\centering
	\renewcommand{\arraystretch}{1.2}
        \scriptsize
        \caption{
            \textbf{\methods can achieve higher consistency across depth than the baseline.} Scores are computed with Format 2 \texttt{[yes, no]}, reported in Appendix \ref{prompts:loco-lms}. \method fine-tuned with on the implication rule and the negation rule achieve best consistency. High sensitivity to prompts should be considered.
        }
        \label{table_eb_1}
        \begin{center}
        \sc
            \begin{tabular}{lrrrrr}
                \toprule 
                         {} & \multicolumn{5}{c}{depth} \\
                         \cmidrule(lr){2-6} 
                         model & 1 & 2 & 3 & 4 & 5 \\
                \midrule
                         LLaMa-2-7b & 1.00 & 0.75 & 0.38 & 0.42 & 0.46 \\
                \midrule
                        LoCo-LLaMa-2-7b (NEG) & \textbf{0.99} & \textbf{0.99} & \textbf{0.99} & \textbf{0.99} & \textbf{0.99} \\
                        LoCo-LLaMa-2-7b (F-IMP) & \textbf{0.99} & \textbf{0.99} & \textbf{0.99} & \textbf{0.99} & \textbf{0.99} \\
                        LoCo-LLaMa-2-7b (Super) & 0.62 & 0.62 & 0.63 & 0.63 & 0.64 \\
                \bottomrule
            \end{tabular}
        \end{center}
\end{table}

\begin{table}[!h]
	\centering
	\renewcommand{\arraystretch}{1.2}
        \scriptsize
        \caption{
            Distribution of answer labels from \methods for different prompt formats on the EntailmentBank test split.
        }
        \label{app:RQ3:eb_dist}
        \begin{center}
        \sc
            \begin{tabular}{lrrrrrr}
                \toprule 
                         {} & \multicolumn{3}{c}{labels: \texttt{[yes, no]}} & \multicolumn{3}{c}{labels: \texttt{[true, false]}} \\
                         \cmidrule(lr){2-4} 
                         \cmidrule(lr){5-7}
                         model & yes & no & invalid & true & false & invalid \\
                \midrule
                         LLaMa-2-7b & 1188 & 6 & 1441 & 615 & 1742 & 278 \\
                \midrule
                         LoCo-LLaMa-2-7b (NEG) & 2538 & 0 & 97 & 940 & 0 & 1695 \\
                         LoCo-LLaMa-2-7b (F-IMP) & 2557 & 0 & 78 & 2441 & 194 & 0 \\
                         LoCo-LLaMa-2-7b (Super) & 2079 & 486 & 70 & 874 & 1756 & 5 \\
                \bottomrule
            \end{tabular}
        \end{center}
\end{table}

\newpage
\section{EntailmentBank}

\begin{figure}[!ht]
    \begin{center}
        \includegraphics[width=0.8\textwidth]{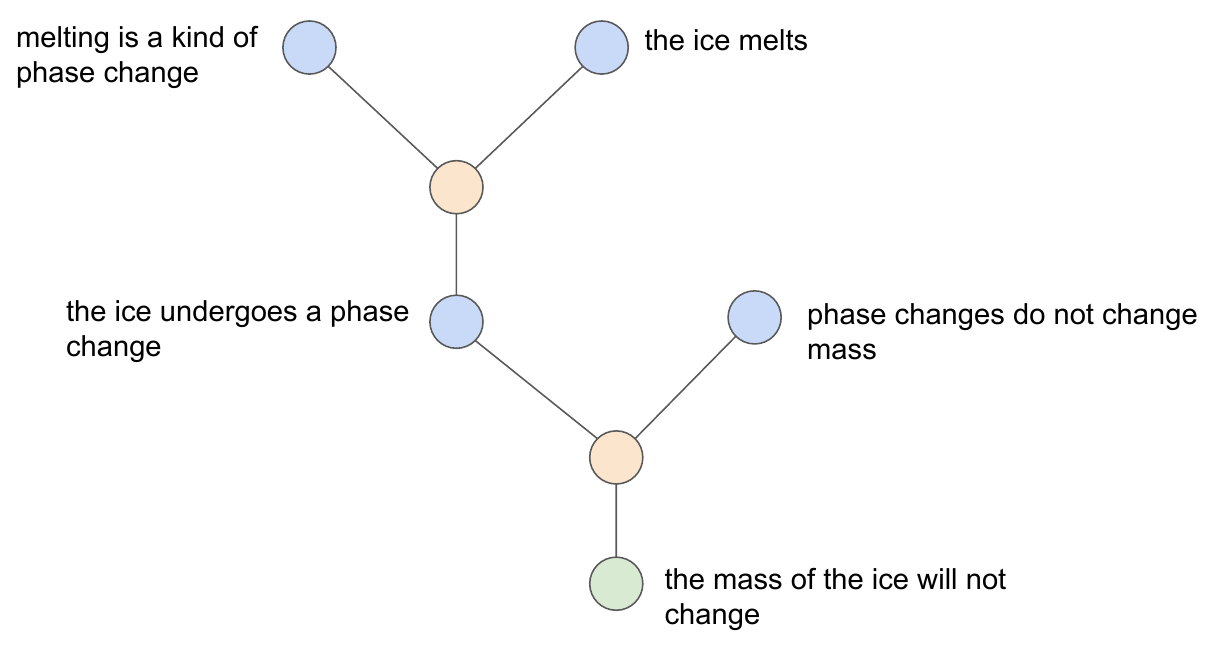}
    \end{center}
    \caption{An illustration of an entailment tree, namely a ``prof", from EntailmentBank \cite{dalvi2022explaining}. Blue nodes are premises in logical conjunction, orange nodes are implications and the green node denote the hypothesis to prove.}
    \label{viz:eb_example0}
\end{figure}

\section{Semantic overlap}
\label{app:semantic-overlap}

We base our measurement for semantic overlap on cosine similarity, widely adopted in literature. We report our results with a note for caution: it is unclear whether embeddings could be similar for the semantic features we are seeking \cite{Steck_2024}, suggesting further research on the topic.  

\subsection{BeliefBank}
\label{sem-op:bb-bb}
We measure the semantic overlap between the training and test distribution by constructing a Representation Dissimilarity Matrix (RDM) of Macaw's embeddings (token average) between training and test entities. The main assumption is that semantically similar subjects may have similar properties, as a proxy for domain knowledge transfer. \\

\begin{figure}[h]
    \begin{center}
        \hspace*{-1in}
        \scalebox{.85}{\includegraphics[width=1.8\textwidth]{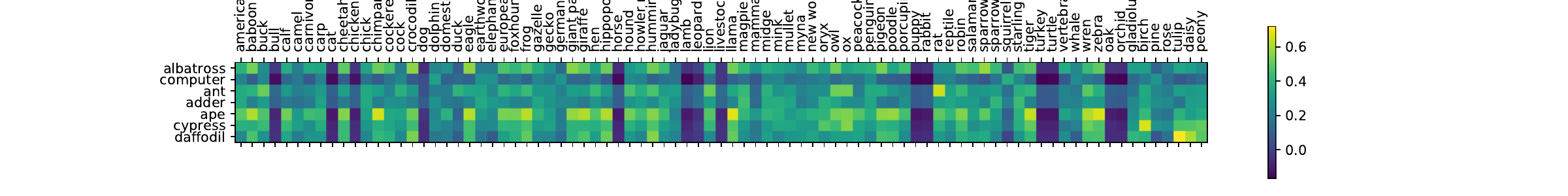}}
    \end{center}
    \caption{Pairwise cosine similarities between entities in the training distribution (calibration, rows) and the test distribution (silver, columns).}
\end{figure}

\subsection{BeliefBank-EntailmentBank}
\label{sem-op:bb-eb}
We consider the training split, namely ``calibration'' in ConCoRD \cite{mitchell2022enhancing}, from BeliefBank \cite{kassner2021beliefbank}, and the test split from EntailmentBank \cite{dalvi2022explaining} to estimate the knowledge that \methods could transfer to entailment trees. We process BeliefBank as a set of $1,072$ facts, while EntailmentBank as a set of $2,635$ facts. Both sets contain statements in natural language that are converted into vector embeddings through encoding with LLaMa-2-7b \cite{touvron2023llama}; the last layer logits are considered and a sentence representation is obtained by averaging across tokens. We consequently compute the pairwise cosine similarities between fact embeddings from both sets. For each fact in BeliefBank, we take the maximum similarity with any fact from EntailmentBank, which should represent the existance of a unit of a similar knowledge between the two datasets. Given the volume of pairwise comparisons, we aggregate the results.

\begin{table}[!t]
        \label{sem-op:bb-eb}
	\centering
	\renewcommand{\arraystretch}{1.2}
        \scriptsize
        \caption{
            Fraction $k$ of facts in BeliefBank with cosine similarity above $t$ with any fact in EntailmentBank, for $t = \{0.80, 0.85, 0.90\}$.
        }
        \begin{center}
        \sc
            \begin{tabular}{lc}
                \toprule 
                         $t$ & $k$ \\
                \midrule
                         0.80 & 0.41 \\
                         0.85 & 0.22 \\
                         0.90 & 0.02 \\
                \bottomrule
            \end{tabular}
        \end{center}
\end{table}

\section{Prompts}

\subsection{Prompts for Macaw-Large}
\label{prompts:macaw}
We query the language model for a belief label about a statement in natural language. We adopt the format:

\begin{tcolorbox}[title=Prompt, colback=gray!10, colframe=gray!30, fonttitle=\bfseries]
    \$answer\$ ; \$mcoptions\$ = (A) <pos\_label> (B) <neg\_label> ; \$question\$ = Is <subject> a <property>?
\end{tcolorbox}

We fix \texttt{<pos\_label> = "Yes."} and \texttt{<neg\_label> = "No."}. We converted the \texttt{(<subject>, <property>)} tuple in natural language with a formatting function provided by Mitchell et al. \citep{mitchell2022enhancing}.

\begin{tcolorbox}[title=Expected answers, colback=gray!10, colframe=gray!30, fonttitle=\bfseries]
    \$answer\$ = <pos\_label> ;
    \$answer\$ = <neg\_label> ;
\end{tcolorbox}

\subsection{Prompts for \methods}
\label{prompts:loco-lms}
We adopt two label sets to make the model less \textit{prompt sensitive}: \\

\textbf{Format 1}: \texttt{[true, false]}
\begin{tcolorbox}[title=Prompt, colback=gray!10, colframe=gray!30, fonttitle=\bfseries]
    You can answer only with "true" or "false". Is the fact true? Fact: <statement> 
\end{tcolorbox}
\begin{tcolorbox}[title=Expected answers, colback=gray!10, colframe=gray!30, fonttitle=\bfseries]
    Answer: true \\
    Answer: false
\end{tcolorbox}
\textbf{Format 2}: \texttt{[yes, no]}
\begin{tcolorbox}[title=Prompt, colback=gray!10, colframe=gray!30, fonttitle=\bfseries]
    You can answer only with "yes" or "no". Is the fact true? Fact: <statement> 
\end{tcolorbox}
\begin{tcolorbox}[title=Expected answers, colback=gray!10, colframe=gray!30, fonttitle=\bfseries]
    Answer: yes \\
    Answer: no
\end{tcolorbox}

\end{document}